\pdfoutput=1
\documentclass[11pt]{article}

\usepackage{EMNLP2023}

\usepackage{times}
\usepackage{latexsym}
\usepackage{graphicx}

\usepackage[whole]{bxcjkjatype}
\usepackage{array}
\usepackage{makecell}
\usepackage{url}
\usepackage{multirow}
\usepackage{pdflscape}
\usepackage{booktabs}
\usepackage{stfloats} 
\usepackage{amsmath}

\usepackage[T1]{fontenc}
\usepackage[utf8]{inputenc}

\usepackage{microtype}

\usepackage{inconsolata}

\title{Specification-Aware Machine Translation and Evaluation for Purpose Alignment}

\author{
  Yoko Kayano\textsuperscript{1,2} \quad
  Saku Sugawara\textsuperscript{1,2} \\
  \textsuperscript{1}The Graduate University for Advanced Studies (SOKENDAI) \\
  \textsuperscript{2}National Institute of Informatics \\
  \texttt{\{yokokayano,saku\}@nii.ac.jp}}

\begin{document}
\maketitle
\begin{abstract}
In professional settings, translation is guided by communicative goals and client needs, often formalized as specifications.
While existing evaluation frameworks acknowledge the importance of such specifications, these specifications are often treated only implicitly in machine translation (MT) research.
Drawing on translation studies, we provide a theoretical rationale for why specifications matter in professional translation, as well as a practical guide to implementing specification-aware MT and evaluation.
Building on this foundation, we apply our framework to the translation of investor relations texts from 33 publicly listed companies.
In our experiment, we compare five translation types, including official human translations and prompt-based outputs from large language models (LLMs), using expert error analysis, user preference rankings, and an automatic metric. 
The results show that LLM translations guided by specifications consistently outperformed official human translations in human evaluations, highlighting a gap between perceived and expected quality.
These findings demonstrate that integrating specifications into MT workflows, with human oversight, can improve translation quality in ways aligned with professional practice.

\end{abstract}

\section{Introduction}
\label{Section:introduction}
High-quality translation in professional settings requires more than a literal rendering of the source text. 
It must also fulfill a communicative purpose, which depends on factors such as the intended function, target audience, and the broader context of the original text \cite{ReissVermeer1984, nord2006}. 
A single source text may yield different translations depending on these factors.%\footnote{We use \textit{source text} for clarity, although \citet{pym2023exploring} proposes \textit{start text} to emphasize the intertextual nature of translation.}

These contextual factors are typically documented as \textit{translation specifications}. 
A specification is a predefined set of conditions that guide the translation process, including purpose, audience, tone, style, and content priorities \cite{ISO17100, JTFevaluation}. 
They help translators make informed decisions and ensure that the translation meets user needs \cite{ReissVermeer1984, nord2006}. 
Without such guidance, translators may struggle to begin the process at all.

Specifications are also essential in translation evaluation. 
Frameworks such as ISO 5060 and the Multidimensional Quality Metrics (MQM) emphasize specification-based assessment \cite{ISO5060, burchardt-2013-multidimensional}. 
\citet{burchardt-2013-multidimensional} state that ``translation quality can only be assessed in terms of whether or not a translation meets specified requirements and meets its communicative purpose.'' 
When specifications are absent or vague, evaluations tend to focus on surface-level features such as lexical accuracy or fluency, rather than on whether the translation achieves its communicative purpose.
This perspective is also central to functionalist theories in translation studies, which hold that quality should be judged by how well a translation fulfills its intended purpose in the target context, rather than by equivalence with the source text \cite{ReissVermeer1984}.

MQM is widely adopted in machine translation (MT) research, including the Conference on Machine Translation (WMT), where it underpins human evaluation \cite{freitag-etal-2021-experts, freitag-etal-2024-llms, zerva-etal-2024-findings}.
Its detailed error typology has contributed to translation evaluation. 
However, specifications are often treated implicitly, and the idea of translation as a goal-oriented process is not fully integrated into MT research. 
As a result, MT outputs often fall short in real-world applications where purpose and audience matter.
This gap is increasingly problematic as industry clients now expect translations to serve specific business objectives \cite{CSA2024}.\footnote{See Appendix~\ref{appendix:CSA} for further discussion.}

\begin{figure}[t]
\includegraphics[width=\linewidth]{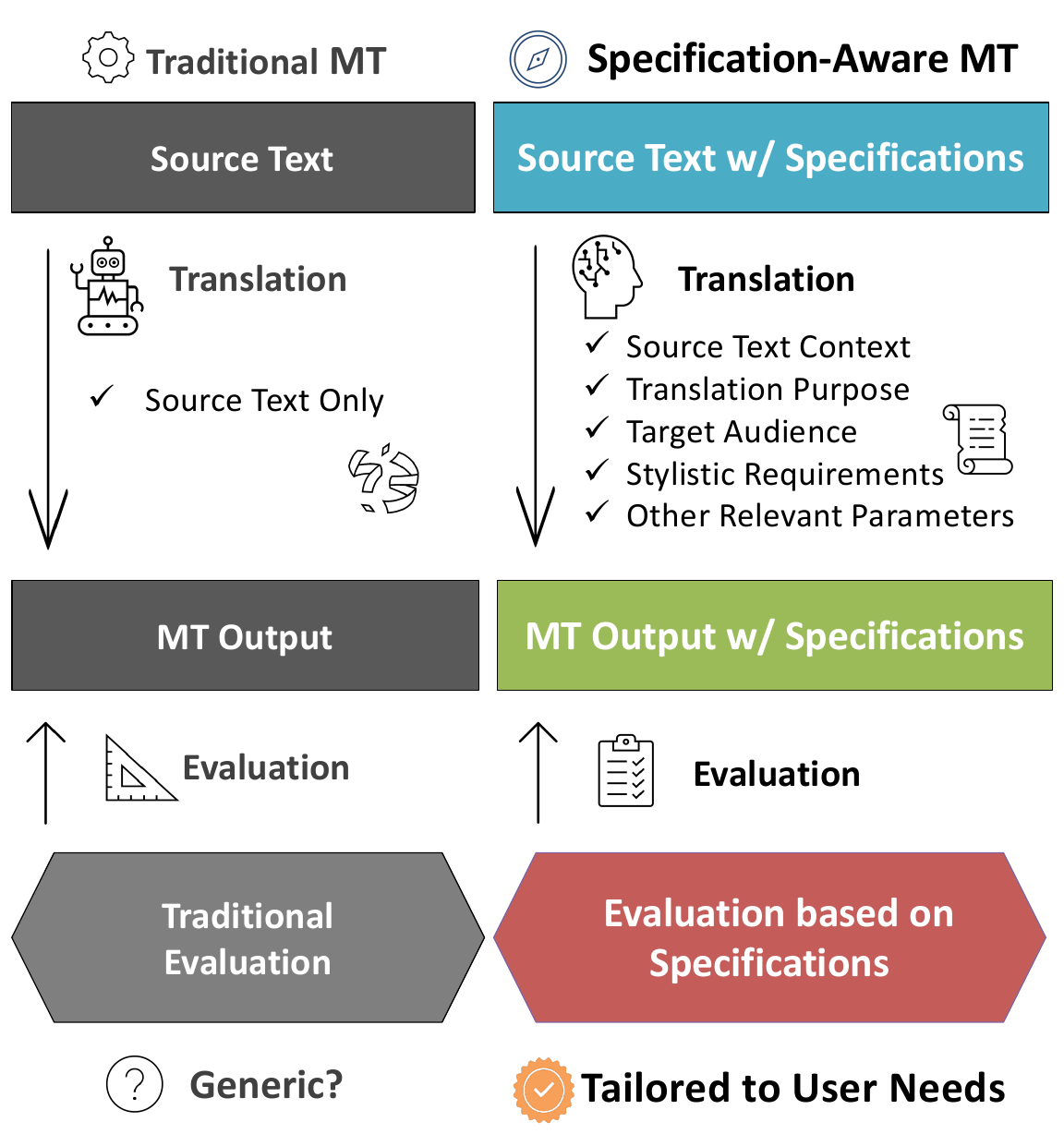}
\centering
\caption{Comparison of traditional MT pipelines, based solely on the source text, and specification-aware pipelines that incorporate contextual and purpose-specific information.}
\label{figure:VS}
\end{figure}

In response, we propose a framework for specification-aware MT and evaluation.
Figure~\ref{figure:VS} outlines this framework, contrasting traditional MT pipelines with our approach, which incorporates contextual information and specification-based evaluation. 
This approach reflects the functionalist view that translation should be guided by purpose, audience, and context. 
Incorporating specifications into MT workflows and evaluation is essential for improving translation quality in professional domains.
To support this claim, we present the theoretical foundation for understanding the role of specifications in Section~\ref{section:theoretical_background}.
We introduce a practical guide based on international standards such as ISO 5060, ISO 17100, MQM, and the JTF guidelines in Section~\ref{section:guide_summary} \cite{ISO5060, ISO17100, MQM, JTFevaluation}.
We apply this framework to a case study in Section~\ref{section:case_study}.

Our case study focuses on investor relations (IR) texts from 33 publicly listed Japanese companies, where alignment with specifications is crucial. 
We compare five English translations: the company's official version, a proprietary MT output, a basic prompt-based LLM output, a prompt-based LLM output using specifications, and a prompt-based LLM post-edit of the MT output.\footnote{Throughout this paper, \textit{LLM} refers to large language models guided by prompt-based customization. We use \textit{post-edit} in a broad sense, including automated revision, and not limited to human editing as defined in ISO 18587 \cite{ISO18587}.}

We evaluate the translations through expert error analysis, user rankings, and a reference-free automatic metric. 
Results show that prompt-based LLM outputs with specifications receive the highest ratings from both experts and users. 
Official translations score lower due to stylistic shortcomings, and conventional MT outputs also underperform. 
In contrast, the automatic metric favors MT output, highlighting a misalignment with human judgment. 
This suggests that specification-aware LLM outputs better fulfill communicative goals, even if not fully reflected in current automatic metrics.\footnote{All translation data and evaluation results are available at \url{https://github.com/nii-cl/Specification_aware_MT}.}

Our contributions are as follows: 
\begin{itemize} 
\item We provide a theoretical foundation for incorporating translation specifications into MT and its evaluation. 
\item We propose a practical guide for applying specifications in MT workflows and test it on IR texts.
\item We demonstrate that specification-aware prompt-based LLMs outperform official human translations in human evaluations, supported by detailed analysis.
\end{itemize}

\section{Related Work}
\label{section:related_work}
\subsection{Customizable Machine Translation}
\label{section:Customizable-MT}
Recent studies have explored how MT can be adapted to specific contexts using external knowledge, prompts, and post-editing beyond the source text.\footnote{For a discussion of prior work on controllable MT in the Statistical Machine Translation (SMT) and NMT eras, see Appendix \ref{appendix:controllable_MT}.}
Though not framed as translation specifications, these efforts share similar goals with ours.

\citet{fujita-2021-attainable} highlights the limits of text-to-text neural machine translations (NMT), emphasizing the need for style guides, terminology, and domain knowledge.
\citet{he-2024-prompting} and \citet{jiao2024gradablechatgpttranslationevaluation} show that prompting GPT-4 with contextual cues and post-editing improves translation quality.
\citet{liu2025effectspromptlengthdomainspecific} further finds that detailed, domain-specific prompts enhance performance in specialized tasks.

For stylistic and functional control, \citet{moslem-etal-2023-adaptive}, \citet{wang-etal-2023-controlling}, and \citet{yamada-2023-optimizing} show that incorporating tone, terminology, and information about the translation's purpose and audience leads to more targeted outputs.
\citet{raunak-etal-2023-leveraging} also demonstrate that post-editing GPT-4 output improves English–Chinese and English–German translation quality.

These studies share motivations with our work on specification-aware translation.
Our study extends these efforts by using real corporate materials and evaluating translation outputs in a practical setting.
We explore the potential of prompt-based LLMs to meet specific professional translation requirements, based on human evaluation.

\subsection{Advances in Translation Evaluation}
Recent work in translation evaluation moves beyond gold references and explores reference-free approaches \cite{blain2023findings, freitag-etal-2023-results, freitag-etal-2024-llms, zerva-etal-2024-findings}. 
Evaluation criteria also expand to include contextual coherence and fine-grained error types.
For document-level automatic evaluation, \citet{vernikos-etal-2022-embarrassingly} improve sentence-level metrics by incorporating context. 
\citet{jiang-etal-2022-blonde} propose BlonDe, a metric that evaluates discourse coherence using span-level F1 scores.
Meta-evaluation, such as that of \citet{moghe-etal-2025-machine}, examine whether metrics can detect diverse error types and highlight their limitations.

MQM-based automatic evaluation has gained traction: GEMBA-MQM \cite{kocmi-federmann-2023-gemba}, AutoMQM \cite{fernandes-etal-2023-devil}, and xCOMET \cite{guerreiro-etal-2024-xcomet} identify error spans and types without language-specific tuning. 
CATER \cite{iida2024caterleveragingllmpioneer} offers reference-free, multi-dimensional evaluation with LLMs, while MQM-APE \cite{lu-etal-2025-mqm} adds automatic post-editing to LLM-based error annotation to focus on quality-improving edits.

Human evaluation also remains essential. 
\citet{lommel-etal-2024-multi} present an MQM scoring framework with calibrated models. 
MQM-Chat \cite{li-etal-2025-mqm} adapts MQM for chatbot, and ESA \cite{kocmi-etal-2024-error} streamlines span-level annotations for non-expert assessments.

Building on these developments, our study incorporates ISO 5060 and MQM principles into a specification-aware evaluation framework.
To better capture translation quality, we assess translations using expert annotators, end-user judgments, and automatic metrics, highlighting both linguistic quality and functional adequacy.\footnote{Appendix~\ref{appendix:related-works-eval} provides more context on discussions of evaluation method reliability and improvement in NLP.}

\subsection{Translation Theory and Machine Translation}
Several studies explore interactions between translation studies and MT research.
\citet{tan2023comparative} apply Skopos-based criteria to compare human and NMT outputs, showing that human translations perform better due to NMT's contextual and lexical limitations.
\citet{liu-etal-2024-evaluation} recommend integrating Skopos theory into human evaluation, while \citet{na2024rethinking} show that theory-informed prompts affect LLM outputs.
\citet{hiebl-gromann-2023-quality} call for a unified concept of translation quality to support collaboration between the fields.

The point raised by \citet{hiebl-gromann-2023-quality} is important: clarifying how the two fields define and evaluate translation quality may help advance both.
To this end, we combine theoretical insights from translation studies with empirical experiments based on real-world workflows, aiming to explore how MT can better address the practical needs of professional translation.
This integration of theory and practice enables a more realistic understanding of MT's role in professional contexts.

\section{Theoretical Background}
\label{section:theoretical_background}
While translation is often seen as producing an equivalent text in another language, the notion of \textit{equivalence} has faced criticism in translation studies since the late 1970s. 
In response, functionalist approaches have gained prominence, viewing translation as a purpose-driven communicative act.
Skopos theory, a widely cited framework, holds that translations should be guided by their purpose.
Based on this view, we argue that translation specifications are essential for developing and evaluating MT systems that meet real-world goals.

We begin with equivalence theory, which frames translation as reproducing the meaning or value of the source text, a view reflected in early MT systems and many current automatic evaluation methods (Section~\ref{section:equivalence}).
We then turn to Skopos theory, a functionalist perspective aligned with our emphasis on translation specifications (Section~\ref{section:skopos}). 
Finally, we discuss how specifications matter not only for translation but also for evaluation (Section~\ref{section:specification}). 

\subsection{Equivalence Theory in Translation Studies and Machine Translation}
\label{section:equivalence}
Equivalence theory \cite{nida1964}, which views translation as reproducing the source text's meaning and value, has long been central to translation studies.\footnote{Equivalence theory includes various perspectives, including formal equivalence, which preserves structure, and dynamic equivalence, which aims for a similar reader response \cite{nida1964, munday2022introducing, pym2023exploring}.}
\citet{dyvik-1992-linguistics} explores this concept in MT, proposing a \textit{situation schema}, an abstract representation that links source and target texts through shared meaning. 
He emphasizes the importance and difficulty of achieving equivalence, even with linguistic theories and technology.

In contrast, \citet{hardmeier-2015-statistical} analyzes how statistical machine translation (SMT) operationalizes equivalence through techniques such as word alignment and domain modeling. 
While SMT reflects equivalence-based assumptions, he argues it oversimplifies translation complexity.

These studies illustrate how earlier MT systems, primarily rule-based and statistical, were shaped by equivalence-oriented thinking.
Although neural networks and deep learning emerged in the mid-2010s \cite{bahdanau2015neural}, earlier systems dominated the field and adhered to formal equivalence.

Even with advances in MT technology, equivalence continues to shape quality evaluation.
Metrics such as BLEU \cite{papineni-etal-2002-bleu} and METEOR \cite{banerjee-lavie-2005-meteor} assess similarity to references, often through n-gram overlap, reflecting a formal equivalence perspective.
Recent model-based metrics like COMET \cite{rei-etal-2020-comet} and BLEURT \cite{sellam-etal-2020-bleurt} seek semantic equivalence, yet rely on source-text alignment.

Since the late 1970s, however, equivalence theory has faced criticism.
\citet{snell-hornby1995} argues that it lacks precision and falsely implies symmetry between languages.
She identifies the 1980s cultural turn as a shift from language-based approaches to views considering sociocultural context and the translator's role \cite{snell-hornby2006}.\footnote{This shift is reflected in major anthologies such as \textit{The Translation Studies Reader} \cite{venuti2021}, whose fourth edition retains only \citet{nida1964} for equivalence theory, omitting figures like \citet{vinayDarbelnet1958} and \citet{catford1965}.}

Although equivalence has become less central in translation studies, it is still prominent in MT practice and evaluation.
While semantic equivalence remains foundational, it does not fully address the diverse purposes and communicative contexts of real-world translation.
Functionalist approaches, such as Skopos theory, offer a useful complement.
Emphasizing the intended function, Skopos theory provides a more practical framework for guiding both human and MT in applied settings.

\subsection{Skopos Theory and the Functional Approach to Translation}
\label{section:skopos}
Skopos theory defines translation not as a linguistic transfer but as an intentional activity to fulfill communicative goals \cite{ReissVermeer1984}.
\citet{nord2006} develops this perspective by introducing translation brief (or specification), a set of instructions outlining the purpose, audience, and conditions for the translation.
She emphasizes that translation decisions are not solely determined by the source text, but by how well it serves its function.

\citet{Gouadec2007} applies the functionalist approach to translation workflows by identifying three criteria: the client's objectives (e.g., increasing sales or enhancing brand image), the user's needs (e.g., clarity in technical documentation), and the relevant usage norms and standards.
As \citet{pym2023exploring} notes, this positions translators as ``language technicians'' who operate within a broader communication strategy, ensuring that the translation fulfills its specific role.

This functionalist perspective, once limited to human translation, may now extend to MT with the emergence of prompt-based LLMs.
Earlier domain-specific MT systems required significant resources, including specialized datasets, expert tuning, and time-consuming model training \cite{saunders2022, wang-etal-2023-controlling}.
In contrast, prompt-based LLMs allow users to specify translation requirements through prompting or fine-tuning, making it easier to adapt translations to their intended purpose (Section~\ref{section:Customizable-MT}).
Although empirical evidence is still emerging, customization is now easier, and the rise of LLMs marks a technological shift that aligns with the functionalist view of translation.

Our study investigates whether and how recent advances in MT, especially prompt-based LLMs, can support a functionalist approach by producing translations aligned with specifications.
This perspective shifts the focus from linguistic equivalence to functional effectiveness and offers insights for improving MT design and evaluation.%\footnote{Skopos theory does not address concept drift, since a \textit{skopos} is defined for a specific project. 
%If a concept's meaning changes, a functionalist approach would require a revised translation brief to match the new context and expectations.}

\begin{table*}[t]
\footnotesize
\centering
\setlength{\tabcolsep}{4pt}
\begin{tabular}{@{}r p{16em} p{30em}@{}}

\toprule
\phantom{1} & \textbf{Parameter} & \textbf{Description} \\
\midrule
1 & Purpose of translation & Communicative goal (e.g., inform, persuade, comply, etc.) \\
2 & Target audience & Intended readers and their language background or expectations \\
3 & Style, register, and tone & Formality, style, and tone appropriate for the target context\\
\midrule
4 & Terminology and reference resources & Use of glossaries, style guides, and prior translations \\
5 & Domain and legal requirements & Industry norms and compliance with relevant laws \\
6 & Cultural adaptation & Adjustments for cultural norms or sensitivities \\
7 & Length and formatting & Constraints on text length, layout, or structure \\
8 & Localization needs & Regional or language variant customization \\
\bottomrule
\end{tabular}
\caption{Translation specification parameters. Items 1–3 are essential; others may vary by project.}
\label{table:specification_parameters}
\end{table*}

\subsection{Why Specifications Matter}
\label{section:specification}
Specifications may include parameters such as purpose, target audience, style, register, domain, timeline, cost, volume, reference materials (e.g., glossaries and style guides), file format, and quality evaluation methods \cite{ISO17100, JTFevaluation}. 
In professional settings, such specifications guide translation decisions and ensure the translation meets client expectations. 
Even if undocumented, essential requirements are typically agreed upon in advance and vary by project. 
For example, legal translations emphasize consistency with terminology and style guides, while marketing texts prioritize creativity and persuasive language.

The importance is also reflected in how translation quality is evaluated.
The MQM framework defines translation quality as follows:
\begin{quote}
A quality translation demonstrates required accuracy and fluency for the audience and purpose and complies with all other negotiated specifications, taking into account end-user needs \cite{melby2012human, burchardt-2013-multidimensional}.
\end{quote}
As explained in the Multi-Range Theory \cite{lommel-etal-2024-multi}, quality evaluation begins with an analysis of project specifications and user needs. 
Evaluators should \textit{select} appropriate error categories and scoring models based on this analysis. 
These ideas are emphasized in the 2024 MQM anniversary paper \cite{lommel-etal-2024-multi}.\footnote{The MQM framework draws on \citet{garvin1984}'s approach to quality. See Appendix~\ref{appendix:Garvin} for further explanation.}

Specifications not only guide translators but also constrain the range of acceptable choices, helping to reduce subjectivity.
As \citet{Gouadec2007} argues, translators are language technicians whose ``plurality is his enemy,'' highlighting the importance of clear instructions \cite{pym2023exploring}. 
This applies equally to evaluation: assessments grounded in specifications are less influenced by personal interpretation. 

Research shows that providing clear criteria and context improves inter-annotator agreement (IAA) \cite{castilho-2021-towards, popovic-2021-agree}. 
The official MQM website also notes that the framework supports standardized, objective evaluation by minimizing subjective judgment \cite{MQM}.
For details on how translation specifications can be incorporated into MQM-based evaluation, see Appendix~\ref{appendix:MQM}.

\section{A Practical Guide for Specification-Aware MT and Evaluation}
\label{section:guide_summary}
We provide a brief overview of our practical framework for integrating translation specifications into both MT workflows and evaluation, where MT is performed using prompt-based LLMs.\footnote{We base our translation and evaluation guidelines on a typical professional workflow and ISO 17100 for translation, and on ISO 5060, JTF guidelines, and MQM for evaluation \cite{ISO17100, ISO5060, JTFevaluation, MQM}.}
A full version is available in Appendix~\ref{appendix:practical-guide}.
\subsection{Specification-Aware Machine Translation with Prompt-Based LLMs}
\paragraph{Step 1: Define Specifications}
Clarify translation requirements. 
These form the basis for both machine output and human review.
Our specification parameters, listed in Table~\ref{table:specification_parameters}, are independently developed based on professional translation practice and informed by existing standards and research \cite{ISO17100, ISO11669_2024, Melby2012_TTTspecs}.

The top three items are essential for all translation projects, regardless of domain or medium.
The remaining items are project-dependent and may be included as needed. 
Additional parameters may be added depending on the context or client requirements.
A brief explanation and examples for each item are provided in Appendix~\ref{appendix:Spec-aware}.

\paragraph{Step 2: Design Instructions}
Specifications should be reflected in prompts or fine-tuning. 
It is important that the instructions also include source text information, target language, and relevant specification parameters, while preventing hallucination and over-generation.
\paragraph{Step 3: Generate and Review}
Use LLMs to generate the translation, followed by human review to ensure the output meets specifications. 
Reviewers make corrections and finalize the translation.

\subsection{Specification-Aware Evaluation}
\label{subsection:spec-eval}
Translation evaluation is not always conducted alongside the translation itself.  
It may be required in various contexts, such as accepting or rejecting a translation, comparing outputs, selecting the best version, ensuring quality in professional workflows, evaluating MT results, or training and certifying translators.
We outline a framework that incorporates both objective and subjective evaluations.

\paragraph{Step 1: Make Specifications Accessible}
Ensure all evaluators have access to the translation specifications. 
If not provided in advance, define them before evaluation begins.
\paragraph{Step 2: Define Error Categories}
Set error categories (e.g., Accuracy, Style, Terminology, etc.) aligned with the specifications. 
Use established frameworks such as MQM and ISO 5060 \cite{MQM, ISO5060}.
\paragraph{Step 3: Weight and Score Errors}
Assign weights to error categories based on project priorities.
Evaluate severity (e.g., minor, major) and calculate a total score using a weighted formula.
\paragraph{Step 4: Add Subjective Evaluation (Optional)}
In addition to error-based scoring, subjective evaluation helps assess whether a translation is appropriate, persuasive, and effective for its intended audience.
Feedback from experts or users can offer insights into clarity, tone, and impact that error metrics alone may overlook.

The following case study demonstrates the application of this practical guide.

\section{A Case Study in Japanese-to-English Translation of Investor Relations Materials}
\label{section:case_study}
We present a case study to show how specification-aware translation can be applied using a prompt-based LLM.
We compare it with human translation and non-prompt-based MT outputs, examining how each is evaluated through both human and automatic methods.
This case study puts into practice the notions discussed earlier in Section~\ref{section:theoretical_background} and assesses the effectiveness of our approach in a real-world Japanese-to-English translation task.

\begin{figure}[t]
\includegraphics[width=0.9\linewidth]{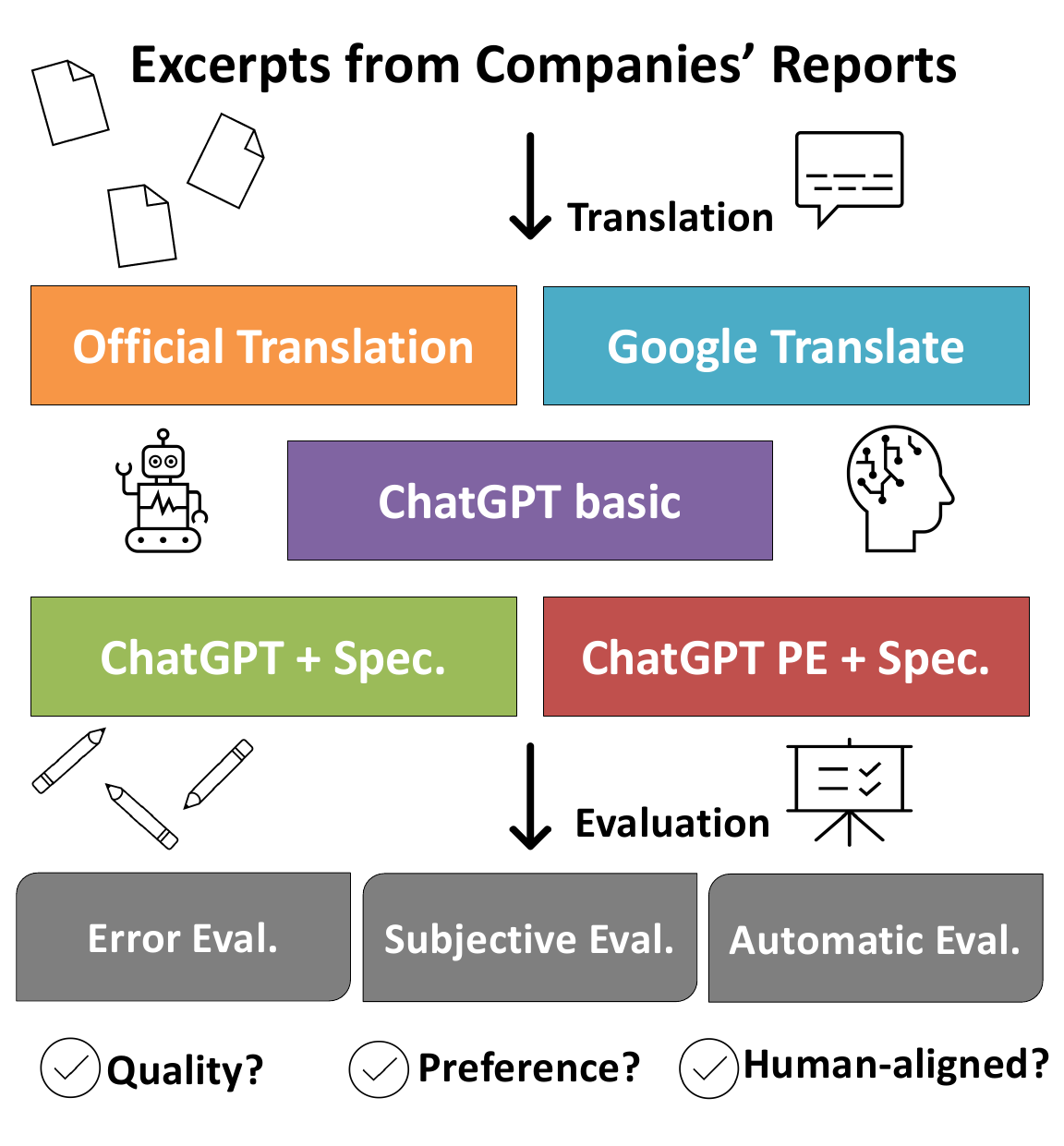}
\centering
\caption{Overview of the case study: Five translation types and their evaluation via expert, user, and automatic methods.}
\label{figure:flowchart}
\end{figure}

Figure~\ref{figure:flowchart} provides an overview of the case study. 
For LLM, we use ChatGPT via its public interface to simulate a scenario in which non-expert users, such as translators or corporate communications personnel, can control translation output through prompting, without needing specialized tools or programming skills.

\subsection{Experimental Setup}
\label{section:experimental_setup}
\subsubsection{Integrated Reports}
\label{subsection:integrated_reports}
We use IR materials excerpted from integrated reports by publicly listed Japanese companies as the source text.
The focus is on Japanese-to-English translation; the rationale for using this language pair is explained in Appendix~\ref{appendix:why-JE}.

Integrated reports combine financial and non-financial information to communicate a company's value creation to investors and other stakeholders.
Although not legally required in Japan, their publication has increased with growing interest in ESG investment.
As of the end of 2023, 1,019 companies issue integrated reports, 70 percent of which also provide English versions \cite{tougouhoukoku}.
Among Prime Market companies, over half publish integrated reports.
We choose integrated reports because they are more structured than websites and more interpretive than financial statements, posing challenges for both human translators and MT systems.

We focus on the corporate philosophy section, typically found at the beginning of these reports.
According to the \textit{Guidance for Collaborative Value Creation 2.0} \cite{kachikyousou}, such statements are central to investor communication and must clearly express a company's unique values.
Translating them requires not only literal accuracy but also clarity, an appropriate corporate tone, and expressions that enhance appeal to stakeholders.

We select one company from each of the 33 industries defined by the Tokyo Stock Exchange \cite{topix33}, prioritizing those with higher market capitalization (27 first-ranked, five second-ranked, and one fourth-ranked).\footnote{The full list appears in Appendix~\ref{appendix:company-list}.}
The extracted sections range from 240 to 927 Japanese characters, with an average of 610.

We manually confirmed alignment with the Japanese source texts and asked companies how their English versions were produced.
Of the 33 companies, 20 responded.
Among these, 15 used only human translation, two combined human and MT, two declined to disclose their method, and one outsourced the work without providing details.
None reported using MT alone, suggesting that human translation remains standard.

\subsubsection{Five Translation Methods} 
\label{subsection:five_translations}

\begin{table*}[!t]
\centering
\fontsize{7.9pt}{10pt}\selectfont
\begin{tabular}{p{3.7em} p{50em}}  
\toprule
\textbf{Type} & \textbf{Translation Output} \\ \midrule
\textbf{Source} & ``ワクワク''は, 人を動かすエネルギー. それは人から人へと伝わり、世界をあかるく元気にする。 \\
\textbf{Official} & ``Waku waku'' is what moves people to push what's possible. It's Japanese for the joy and excitement
of discovering the unknown.
And when passed from person to person, becomes a force that creates a brighter world, united in wonder.\\
\textbf{Google} & ``Excitement'' is the energy that moves people. It spreads from person to person, making the world brighter and more energetic.  \\
\textbf{GPT-b} & ``Excitement'' is the energy that moves people. It spreads from person to person, bringing brightness and vitality to the world. \\
\textbf{GPT+Sp} & Excitement is the energy that moves people. It spreads from person to person, brightening and invigorating the world. \\
\textbf{PE+Sp} & ``Excitement'' is the spark that moves people, spreading from one person to another, brightening and energizing the world.\\
\bottomrule
\end{tabular}
\caption{Differences in translations (All Nippon Airways Co., Ltd.)}
\label{table:translations}
\end{table*}

To compare translation quality and effectiveness, we prepare five versions using different methods.
The official translation consists of excerpts from English versions of integrated reports published by the companies.

We then create four MT-based versions:
\begin{itemize}
\item \textbf{Google Translate}: raw output from Google Translate
\item \textbf{ChatGPT basic}: ChatGPT with a minimal prompt
\item \textbf{ChatGPT + Spec}: ChatGPT with specifications
\item \textbf{ChatGPT PE + Spec}: Google Translate post-edited by ChatGPT with specifications
\end{itemize}
To ensure consistency, we use the first output for all versions.
All ChatGPT translations are generated using ChatGPT-4o.

For the specification-aware methods, we provide prompts that reflect key information such as source text context, intended purpose (e.g., appealing to global investors), target audience, and stylistic tone.
The full prompt is shown in Appendix~\ref{appendix:prompts}.

Using these methods, we generate five translations for each of the 33 companies.
Manual review indicates that all versions maintain overall meaning without serious accuracy errors.
However, we observe a recurring issue in ChatGPT translations: kanji misinterpretation.
For example, \textit{文殊院旨意書} (Monjuin Shiigaki) is rendered incorrectly as \textit{Monjuin Shiisho}, lacking accurate transliteration.
This suggests that ChatGPT struggles with domain-specific terminology and proper nouns.

Table~\ref{table:translations} shows translations of a corporate philosophy excerpt from All Nippon Airways Co., Ltd.'s integrated report.
The official translation contains a grammatical error (``And when passed... becomes...'') and phrases that may be unclear or unnatural (``to push what's possible'').
It also gives an extended explanation of \textit{waku waku}.
The Google translation is grammatically correct but closely mirrors the source, resulting in a literal tone and basic vocabulary.
The ChatGPT basic version improves fluency and uses slightly richer expressions (``vitality''), but its tone and structure remain similar to the Google version.
The ChatGPT version with specifications uses more active verbs and parallel phrasing (``brightening and invigorating''), resulting in smoother rhythm and tone.
The post-edited version with specifications introduces vocabulary like ``spark'' and ``energizing,'' while preserving the original meaning and structure.

These examples show that each method yields distinct results and that adding specifications to ChatGPT prompts may encourage more purposeful and expressive language.
Appendix~\ref{appendix:example_ANA} provides a comparative analysis of a longer excerpt from the same source, focusing on linguistic and stylistic differences.

\subsection{Human Evaluation}
\label{section:case_human_eval}
After preparing translations for all 33 companies, we conduct two human evaluations of the five translation methods: expert error evaluation and subjective evaluation. 

\subsubsection{Error Evaluation}
\label{subsection:case_error_eval}
We conduct an error-based evaluation using the specification-aware framework introduced in Section~\ref{subsection:spec-eval} and detailed further in Appendix~\ref{appendix:human_error}.
This evaluation focuses on three core categories: Accuracy, Linguistic Conventions, and Style.
Other categories defined in the MQM framework, such as Design and Markup, are excluded as they are not applicable to the scope of this study.

All category definitions are based on the MQM standard and were provided to the evaluators to ensure consistency and shared understanding \cite{MQM}.
See Appendix~\ref{appendix:error-typology} for the detailed error categories used in the annotation.

Given the importance of stylistic quality in IR materials, we include four subtypes under Style:
(1) Language register mismatch, (2) Awkward style, (3) Unidiomatic expressions, and (4) Inconsistent style.
These errors do not hinder comprehension but result in unnatural English that may reduce clarity and impact.
Subtypes help clarify scope, but annotators classify errors only at the main category level to reduce cognitive burden.
Error categories are weighted based on JTF guidelines: Accuracy (0.7), Linguistic Conventions (0.8), and Style (1.5), averaging to 1.0 overall \cite{JTFevaluation}.
We do not apply severity levels, as the texts do not involve high-stakes content such as financial figures.

%Two professional evaluators, each a professional translator or an expert in linguistics and culture, and a native speaker of the target language (English), 
Two professional evaluators, each either a professional translator or an expert in linguistics and culture, bilingual in Japanese and English, and with English as their first language, are recruited via Prolific.\footnote{\url{https://www.prolific.com}}
They receive the Japanese source text, translation specifications, five anonymized English translations, an error typology table with definitions, and sample annotations.
They identify errors, assign them to one of the three categories, mark their locations, and record error counts. 
Each evaluator is compensated £40 for approximately 270 minutes of work.
Only two out of 24 recruited participants completed the task, highlighting the practical difficulty of securing qualified evaluators and the cognitive demands of error annotation, as noted in prior research  \cite{kocmi-etal-2024-error, zouhar2025ai}.

\begin{table}[t]
\centering \small
\setlength{\tabcolsep}{2.6pt} 
\begin{tabular}{lccccc}
\toprule 
\textbf{Type} & \textbf{Official} & \textbf{Google} & \textbf{GPT-b} & \textbf{GPT+Sp} & \textbf{GPT PE+Sp} \\
\midrule
Eval. 1 & \textit{2.60} & 1.82 & 1.04 & 0.70 & \textbf{0.38} \\
Eval. 2 & \textit{3.01} & 2.29 & 1.28 & 1.29 & \textbf{1.03} \\
\bottomrule
\end{tabular}
\caption{
Weighted error scores averaged across 33 companies. Lower scores indicate fewer errors and higher translation quality. 
}
\label{table:error_scores}
\end{table}

Table~\ref{table:error_scores} presents the evaluation results.
ChatGPT PE + Spec receives the lowest error score (highest quality), followed by ChatGPT + Spec, ChatGPT basic, and Google Translate.
The official translation ranks lowest, with particularly frequent Style errors, which will be discussed in Section~\ref{subsection:case_subjective_eval_analysis}.
These findings suggest that LLM-based translations guided by specifications can outperform human translations in this context, challenging the assumption that human translations should serve as the default gold standard in MT evaluation.

We also assessed inter-annotator reliability by calculating the correlation between the error scores assigned by the two evaluators. 
Pearson's correlation is very high ($r$ = 0.985 and $p$ = 0.0021), while Spearman's rank correlation is also strong ($\rho$ = 0.90 and $p$ = 0.037), indicating statistically significant agreement.
Nonetheless, we observe inconsistencies: the same expression was sometimes marked as an error in one translation but not in another by the same evaluator.
This indicates the inherent difficulty of ensuring consistency in error-based evaluation, even among professionals.
The low completion rate suggests that translation evaluation is time-consuming, cognitively demanding, and difficult to delegate, as it requires a high level of expertise.
Our evaluation process incidentally reflected these challenges in practice.

\subsubsection{Subjective Evaluation} 
\label{subsection:case_subjective_eval}
%To supplement the expert-based error evaluation, we conduct a subjective evaluation to capture how translations are perceived by typical end users.
To understand how translations are perceived by intended end users, we conduct a subjective evaluation alongside expert-based error analysis.
As discussed in Section~\ref{subsection:spec-eval} and detailed in Appendix~\ref{appendix:human_subjective}, combining error-based and subjective evaluation is useful not only when qualified annotators are limited, but also when end-user perspectives take precedence.
For texts like integrated reports, which aim to build trust and attract investment, reader impression may matter more than linguistic accuracy, making subjective feedback particularly valuable.

Subjective evaluation is generally divided into expert and end-user perspectives \cite{JTFevaluation}.
Since integrated reports target investors, we adopt an end-user perspective.
We recruited eighteen native English speakers via Prolific. 
Participants are compensated £13.50 for approximately 90 minutes of work.
Seventeen hold degrees in fields such as accounting or finance. 
One participant is a translator and linguistic expert who also took part in the error evaluation.
Each evaluator receives the translation specifications and five English translations, without knowing the translation methods or having access to the Japanese source. 
They are asked to rank the translations based on overall appeal, defined as clarity, readability, word choice, and company presentation.

\begin{figure}[t]
\includegraphics[width=\linewidth]{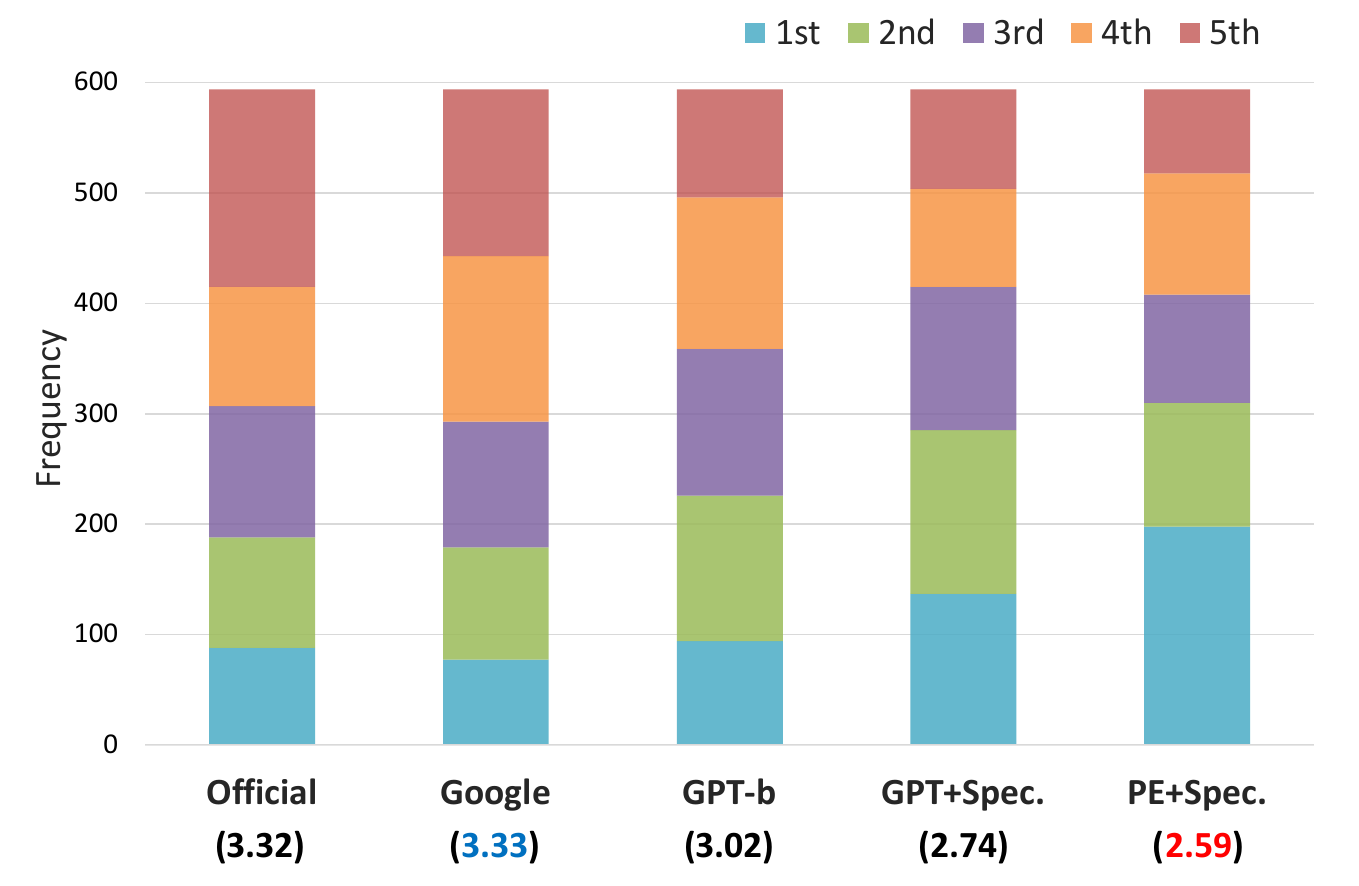}
\centering
\caption{Ranking counts from the subjective evaluation. The x-axis shows translation methods, and the y-axis shows frequency. Each bar is stacked by rank (1st to 5th). Numbers in parentheses indicate mean rankings; lower values reflect higher preference.}
\label{figure:ranking_count}
\end{figure}

Figure~\ref{figure:ranking_count} shows the distribution of rankings across translation types.
ChatGPT PE + Spec is most often ranked 1st, whereas the official translation is most often ranked 5th.
Google Translate is the least often ranked 1st, while the three ChatGPT-based translations are less often ranked last.
Numbers in parentheses indicate mean rankings: ChatGPT PE + Spec has the best (lowest) average ranking, followed by ChatGPT + Spec, ChatGPT basic, and the official translation. Google Translate ranks lowest overall.

To assess significance, we conduct Wilcoxon signed-rank tests on all ten translation pairs, reporting the test statistic ($W$), $p$-values, and effect sizes ($r$) in Table~\ref{table:wilcoxon}.
A $p$-value below 0.05 is considered significant; effect sizes are interpreted as small ($r = 0.1$), medium ($r = 0.3$), or large ($r = 0.5$).
All comparisons are significant except Official vs. Google Translate and ChatGPT + Spec vs. ChatGPT PE + Spec.
Pairs with ChatGPT PE + Spec (vs. Official and Google Translate) show the strongest effects, with $p \approx 0$ and medium effect sizes.

These results show that ChatGPT PE + Spec is consistently preferred, in line with error-based evaluation findings.
The low ranking of the official translation is notable, despite its presumed status as the gold standard.
However, human translations often vary more than MT, depending on translator performance \cite{freitag-etal-2023-results, PrietoRamos02042024, VolzThiessen2024}.
%This variation is reflected in our data: the official translation is most frequently ranked last, though it is more often placed 3rd than 4th.
%This variation is apparent in the wide distribution of rankings for the official translation. 
%Despite being ranked last most often, its perceived quality is not uniform among evaluators.
As a result, while the official translation was most frequently ranked in the lowest position, the number of times it was placed second, third, or fourth did not differ significantly. 
Moreover, it was ranked first more often than Google Translate and not far behind ChatGPT basic.

\begin{table}[t]
\centering \footnotesize
\setlength{\tabcolsep}{3pt}
\begin{tabular}{lcccc}
\toprule
\textbf{Pair} & \textbf{$W$} & \textbf{$Z$} & \textbf{$p$} & \textbf{$r$} \\
\midrule
Off. vs Ggl & 88018 & 0.081 & 0.9341 & 0.003 \\
Off. vs GPT-b & 74913 & 3.213 & 0.00113 & 0.132 \\
\textbf{Off. vs GPT+Sp} & \textbf{63954} & \textbf{5.832} & \textbf{$p < 0.00001$} & \textbf{0.239} \\
\textbf{Off. vs PE+Sp} & \textbf{59011} & \textbf{7.013} & \textbf{$p< 0.00001$} & \textbf{0.288} \\
Ggl vs GPT-b & 73843 & 3.469 & 0.00043 & 0.142 \\
\textbf{Ggl vs GPT+Sp} & \textbf{62409} & \textbf{6.201} & \textbf{$p< 0.00001$} & \textbf{0.254} \\
\textbf{Ggl vs PE+Sp} & \textbf{57082.5} & \textbf{7.474} & \textbf{$p < 0.00001$} & \textbf{0.307} \\
GPT-b vs GPT+Sp & 75367 & 3.104 & 0.00162 & 0.127 \\
GPT-b vs PE+Sp & 68326.5 & 4.787 & $p< 0.00001$ & 0.196 \\
GPT+Sp vs PE+Sp & 81396.5 & 1.664 & 0.0903 & 0.068 \\
\bottomrule
\end{tabular}
\caption{Wilcoxon signed-rank test for translation pairs. A significant difference is defined as $p < 0.05$. Effect size ($r$) is interpreted as small (0.1), medium (0.3), and large (0.5).}
\label{table:wilcoxon}
\end{table}

\subsubsection{Qualitative Analysis} 
\label{subsection:case_subjective_eval_analysis}
 To gain insight into the stylistic and structural differences across translation types, we examine their sentence structure, focusing on relative clauses and clausal coordination.
Our analysis shows that Google Translate and the official translations tend to use these forms more frequently, potentially reflecting source-language influence.
Japanese allows for long, additive sentence constructions, which can lead to overuse of relative clauses or clausal coordination when translated too literally into English.
Such structures may reduce readability, especially in English writing that values clarity and conciseness.
Further details are provided in Appendix~\ref{appendix:syntax}.

We also analyze excerpts from the official translations and find recurring issues in grammar, style, and semantic clarity. 
For example, the expression ``offering both a multitude of choices'' contains a semantic mismatch between ``both'' and ``multitude.'' 
Other examples involved unidiomatic phrasing, sentence fragments, and inconsistent style. 
For detailed examples and qualitative error analysis, see Appendix~\ref{appendix:official_examples}. 
These problems suggest that the low rating of the official translation may not stem from a lack of specifications but from variation in translator skill or mismatches with task requirements.

As mentioned in Section~\ref{subsection:case_subjective_eval}, human translations often vary in quality due to individual differences \cite{freitag-etal-2023-results, PrietoRamos02042024, VolzThiessen2024}. 
Combined with Japan's shortage of high-proficiency English translators (Appendix~\ref{appendix:why-JE}), this may explain the observed results. 
By contrast, ChatGPT-based translations guided by specifications performed consistently well, suggesting their potential as a viable complement to traditional workflows.

\subsection{Automatic Evaluation}
\label{section:auto_eval}
We examine whether a reference-free automatic metric can capture differences in translation quality across specification and method types, compared to human judgment.
To this end, we use COMETKiwi, a reference-free metric with the highest correlation to human evaluations in the WMT23 Metrics Shared Task \cite{rei-etal-2022-cometkiwi,freitag-etal-2023-results}.\footnote{We use the model \texttt{wmt22-cometkiwi-da}, also adopted as the WMT24 baseline for reference-free evaluation \cite{freitag-etal-2024-llms}.}
We adopt a reference-free approach because the official translations, typically used as references, are themselves part of the evaluation as one of the five translation types.

\begin{table}[t]
\centering \small
\setlength{\tabcolsep}{2.6pt}
\begin{tabular}{lccccc}
\toprule
\textbf{Type} & \textbf{Official} & \textbf{Google} & \textbf{GPT} & \textbf{GPT+Sp} & \textbf{GPT PE+Sp} \\
\midrule
Mean & 
\textit{0.783} & 
\textbf{0.830} & 
0.822 & 
0.821 & 
0.810 \\
SD &
0.043 &
0.031 &
0.039 &
0.033 &
0.037 \\
\bottomrule
\end{tabular}
\caption{Mean COMETKiwi scores and standard deviations for each translation type.}
\label{table:cometkiwi}
\end{table}

Table~\ref{table:cometkiwi} shows scores from 0 to 1, with higher values indicating better quality.
Low standard deviations suggest internal consistency, though overlapping ranges point to limited differences between types.
Unlike the human rankings, COMETKiwi assigns the highest score to Google Translate and the lowest to the official translation.
This divergence likely reflects differences in what COMET values, specifically literal fidelity and lexical similarity, as opposed to the more context-sensitive and stylistic qualities emphasized in our evaluation \cite{rei-etal-2022-cometkiwi}.

Although the official translation appears to preserve source-like structures such as relative clauses and clausal coordination (Section~\ref{subsection:case_subjective_eval_analysis}, Appendix~\ref{appendix:syntax}),
its low score may be partly explained by a few explanatory additions not present in the source, intended to assist international readers.
Such additions may reduce source alignment and result in lower automatic scores.

ChatGPT PE + Spec scores slightly below Google Translate, though the difference is small.
This may reflect Google Translate's more literal style, while ChatGPT PE + Spec balances fidelity and fluency, resulting in higher subjective appeal despite a lower COMET score.
ChatGPT translations, particularly those guided by specifications, prioritize clarity and appeal over strict lexical matching, which COMET may not fully capture.

Although COMET metrics are known to struggle with numbers and named entities \cite{amrhein-sennrich-2022-identifying}, our manual check found no significant errors in these areas, suggesting they did not affect the results.

As MT evaluation increasingly considers contextual and communicative goals, it is vital to develop automatic metrics that better capture functional aspects of translation quality, such as how well a translation fulfills its purpose in context.

\section{Conclusion}
\label{section:conclusion}
We demonstrate that translation specifications can improve MT quality and enable more targeted evaluation.
We provide a theoretical rationale for the importance of specifications, drawing on Skopos theory to support a functionalist perspective.
Based on this foundation, we outline a practical guide for specification-aware MT using LLMs, including prompt design, generation, and both error-based and subjective evaluation.

In our case study, LLM outputs guided by specifications received higher ratings than official translations, Google Translate, or unguided LLM outputs.
Although COMET scores favored Google Translate, they diverged from human evaluations.
These findings suggest that specifications help LLMs produce more contextually appropriate translations that better align with communicative goals.
The gap between human and automatic evaluations highlights the limitations of current metrics in capturing functional adequacy.
Through this work, we demonstrate the potential of specification-aware MT for professional, real-world use cases.

\section*{Limitations}
\label{section:limitations}
First, we use only a single LLM, ChatGPT.
While its outputs are generally well-received, it occasionally introduces information not present in the source text.
This highlights the importance of careful prompt design and human oversight, as is standard in professional translation workflows.
Evaluating other LLMs remains an important area for future research to assess whether the findings generalize across models.

Second, our dataset consists of corporate philosophy statements from 33 Japanese companies, focusing solely on the Japanese-to-English language pair. 
While this allowed for a focused case study, broader validation will require larger datasets covering more diverse domains (e.g., legal and medical) and content types (e.g., marketing and technical manuals), as well as other language pairs.

Finally, our human evaluation process highlighted the difficulty of securing qualified annotators.
Both the error analysis and the subjective evaluation were conducted through crowd-sourcing, and the compensation was set above the standard rates of that framework. 
For error analysis, which requires more specialized expertise, an alternative approach could have been to recruit evaluators through a more specialized platform and set the compensation accordingly.
Such difficulties in recruiting and compensating qualified annotators emphasize the need to develop an automated and specification-based evaluation model. 
Future work could explore the \textit{LLM as a Judge} \cite{NEURIPS2023_91f18a12, kocmi-federmann-2023-large, feng-etal-2025-mad, Gunathilaka2025}, where an LLM evaluates outputs based on the same detailed specifications provided to human experts, potentially offering a scalable alternative to manual annotation.

\section*{Acknowledgments}
\label{section:acknowledgments}
We sincerely thank the anonymous reviewers for their insightful and constructive comments, which helped us improve this paper.
This work was supported by JST FOREST Grant Number JPMJFR232R.

\bibliography{anthology,custom}
\bibliographystyle{acl_natbib}

\appendix
 
\section{Industry Trends from CSA Research}
\label{appendix:CSA}
According to \citet{CSA2024} from CSA Research, an independent research firm specializing in the language services industry, language service providers (LSPs) that rely heavily on traditional human translation are experiencing declining performance. 
At the same time, while the growing demand for translation has outpaced the capacity of human translators, the increased use of MT alone has not led to sustainable growth \cite{CSA2024}.

The report encourages LSPs to shift their focus from merely producing high-quality translations to delivering greater value, such as providing cultural adaptation, adapting content for specific audiences, and training and customizing LLMs for domain-specific communication. 
It clearly states: ``LSPs must focus on messaging that resonates with enterprise goals, and demonstrate that they use technology to achieve them \cite{CSA2024}.''

In response to these challenges, our study proposes a framework for specification-aware MT and evaluation. 

\section{Historical Context of Controllable MT}
\label{appendix:controllable_MT}
As noted in Section \ref{section:Customizable-MT}, the goal of tailoring translation output is not new. 
In the eras of SMT and NMT, significant research focused on incorporating external knowledge to control specific aspects of translation.

For example, a major line of work involved leveraging existing human translations to improve consistency. 
This began with the convergence of Translation Memories and SMT \cite{koehn-senellart-2010-convergence} and was later adapted to NMT, such as through neural fuzzy repair mechanisms \cite{bulte-tezcan-2019-neural}. 
Other approaches focused on controlling discrete linguistic features, including methods to enforce terminology constraints during NMT decoding \cite{dinu-etal-2019-training} and to manage stylistic aspects such as formality \cite{niu2020controlling}.

While these approaches provided powerful control over discrete phenomena, they often required specialized data preparation or model retraining. 
Our work builds upon this tradition but explores how prompt-based LLMs can manage a broader set of communicative specifications in a more flexible manner.

\section{Reliability of Evaluation Methods}
\label{appendix:related-works-eval}
Recent research has highlighted the need for more robust and transparent evaluation methods across NLP tasks, including but not limited to MT. 
This includes a growing interest in developing evaluation frameworks that are comprehensive, detailed, and interpretable.
For instance, \citet{nimah-etal-2023-nlg} propose the \textit{Metric Preference Checklist}, an analytical framework that evaluates automatic natural language generation (NLG) metrics from five distinct perspectives, providing a more multifaceted evaluation of their alignment with human judgments.

In contrast, \citet{xiao-etal-2023-evaluating-evaluation} point out that most research focuses only on how well metrics correlate with human ratings, often overlooking the reliability and measurement error of the metrics themselves. 
They argue that concepts from Measurement Theory, used in educational and psychological testing, should be applied to NLG evaluation to better assess the reliability and validity of evaluation metrics.

\citet{gehrmann2023repairing} outline a long-term agenda for improving NLG evaluation, including robust human evaluation protocols and the development of metrics that go beyond surface-level overlap.
Meanwhile, \citet{ruan-etal-2024-defining} emphasize the low reliability of human evaluation guidelines in NLG, showing that only 29.84 percent of 3,233 papers in major NLP conferences shared their guidelines, and 77.09 percent of those contained some kind of vulnerability.
They propose principles for more reliable guideline design and introduce a method using LLMs to detect guideline flaws.

These concerns are relevant to evaluation in MT. 
If the evaluation procedures are unclear, the reliability of the results cannot be ensured.
In our study, we develop guidelines for both translation and evaluation and describe the experimental procedure based on these guidelines.
In addition, instead of simply reporting the results, we offer analysis and possible interpretations for each finding to improve clarity and transparency.

\section{Garvin's Approach to Understanding Quality}
\label{appendix:Garvin}
Since the MQM framework is grounded in \citet{garvin1984}'s approach to quality, we provide a brief overview of his perspectives.

\citet{garvin1984}, a prominent scholar in quality management, introduces five approaches to understanding quality: transcendent, product-based, user-based, manufacturing-based, and value-based.  
Among these, the manufacturing-based approach defines quality as meeting pre-set specifications, and the user-based approach emphasizes satisfying user needs.  

\citet{fields2014whatisquality} discuss the importance of incorporating \citet{garvin1984}'s approach into translation quality assessment. 
Although the definition of translation quality is debated and \citet{fields2014whatisquality} disagree on some points, they generally agree that the production-based approach (originally ``manufacturing-based'' in Garvin's words), evaluating translation according to specifications, is important.

\section{Incorporating Specifications in MT and its Evaluation: The Case of MQM}
\label{appendix:MQM}
Current research using MQM in MT evaluation often focuses on detailed error assessment.  
For example, the MQM framework has been used in shared tasks at the Conference on Machine Translation (WMT), such as the General Translation Task, the Metrics Task, and the Quality Estimation Shared Task. 

However, specifications are not explicitly addressed in these tasks.  
This is because the purpose of the General Translation Task is to evaluate general MT capabilities, which may not require specific requirements, such as the purpose of the translation or the target audience, to translate the source text.  
This kind of general MT capability can be useful when users simply want to gain a general understanding of foreign content. 
However, if the goal is to communicate clearly in the target language, the translation must convey the message in a way that reflects its purpose and suits the intended reader.
This cannot be done without specifications.

\citet{freitag-etal-2021-experts} report that in MQM evaluations of English-to-German and Chinese-to-English translations, approximately 80 percent of errors fall into the accuracy and fluency categories, with accuracy-related mistranslations being the most common.  
While accuracy and fluency are relatively straightforward to assess, other error categories may be more difficult to judge without specifications. 
For example, the definition of the \textit{Style} error is simply ``Translation has stylistic problems'' \cite{freitag-etal-2021-experts}. 
While this is distinct from \textit{Fluency-grammar} errors, it may be difficult for evaluators to identify a stylistic issue if the purpose of the translation is not known. 

In specification-based evaluation, if the purpose of the translation is to convey the cultural otherness of the source text and the specified style is a literal translation that closely follows the original, the translation should adhere to that style.
In this case, the translation is expected to preserve the expressions and cultural markers of the source to maintain the visibility of the translation.
In other words, even a natural and fluent translation may be considered an error if it minimizes the sense of translation under such specifications.

As mentioned earlier, specifications can also serve as a guide for evaluators, helping to reduce subjectivity in the evaluation process. 
In this sense, the detailed error types in MQM could be applied more effectively when used in conjunction with detailed specifications.

Several recent MQM-based approaches, such as GEMBA-MQM \cite{kocmi-federmann-2023-gemba}, Auto-MQM \cite{fernandes-etal-2023-devil}, xCOMET \cite{guerreiro-etal-2024-xcomet}, and MQM-APE \cite{lu-etal-2025-mqm}, have laid important groundwork for the automation of fine-grained evaluation. 
However, translation specifications are not the central focus of these approaches.

While human evaluation approaches like MQM-Chat adapt MQM error categories to specific settings (e.g., chatbot applications), they do so by simply modifying the original MQM typology \cite{li-etal-2025-mqm}.
Their approach could be further extended to integrate detailed specifications.
For example, it could require translations to preserve source-text ambiguity as a stylistic feature or to handle internet slang accurately as part of terminology, in line with MQM's original design.

These considerations highlight that MQM, while widely used, can be made more effective when applied in combination with translation specifications.

\section{A Practical Guide for Specification-Aware Machine Translation and Evaluation}
\label{appendix:practical-guide}
We present guidelines for specification-aware translation and evaluation, drawing on the translation process in a typical professional workflow and the ``Production Process'' outlined in ISO 17100—Requirements for Translation Services. 
This standard states that translation should be ``in accordance with the purpose of the translation project, including the linguistic conventions of the target language and relevant project specifications'' \cite{ISO17100}.

Our evaluation method draws on ISO 5060, the JTF guidelines, and the MQM framework \cite{ISO5060, JTFevaluation, MQM}.\footnote{The European Commission's Directorate-General for Translation also provides quality evaluation guidelines based on a ``fit for purpose'' approach \cite{strandvik2017DGT}. 
Translations are assessed using error categories (e.g., Accuracy, Terminology, Linguistic Conventions, Style, and Formatting) with severity levels to calculate a quality score \cite{DGT2015}.}
Our approach combines these frameworks to make use of their different strengths in evaluation.
MQM provides detailed error types for fine-grained analysis, while ISO 5060 and the JTF guidelines allow for weighted error categories based on specifications.\footnote{The error classification used in ISO 5060 is based on the MQM framework \cite{ISO5060}.}
The JTF guidelines also emphasize the role of subjective evaluation \cite{JTFevaluation}.
These evaluation guidelines are designed for use with both human and MT and assume that human evaluators will assess the output.  

We first explain the method of specification-aware translation with prompt-based LLMs (Appendix~\ref{appendix:Spec-aware}), followed by a description of the human error analysis procedure (Appendix~\ref{appendix:human_error}).
It also addresses subjective evaluation, which provides a different perspective from error analysis for assessing whether the translation fulfills its intended purpose (Section~\ref{appendix:human_subjective}).

\subsection{Specification-aware Machine Translations with Prompt-Based LLMs}
\label{appendix:Spec-aware}
\paragraph{Step 1: Define Translation Specifications}
Define detailed translation specifications. 
When working with clients or translation teams, all stakeholders should reach an agreement on the specifications. 
The following are examples corresponding to the parameters listed in Table~\ref{table:specification_parameters}, using a single translation project as context: the English translation of an integrated report published by a publicly listed Japanese company.

\begin{enumerate}
\item \textbf{Purpose of translation:} The translation's communicative goal (e.g., to inform, persuade, promote, or comply).

Example: The goal is to attract foreign institutional investors. Therefore, the translation must emphasize growth potential and sustainability strategies in persuasive, investor-friendly language.

\item \textbf{Target audience:} The intended readers and their background knowledge, expectations, or needs.

Example: The audience consists of non-Japanese institutional investors who may not be familiar with Japanese corporate structures. Key terms may require explanatory phrasing.

\item \textbf{Style, register and tone:} The desired level of formality, stylistic conventions, and voice appropriate to the target context.

Example: A moderately formal and confident tone is preferred, neither overly technical nor overly casual, consistent with ESG reports by global competitors.

\item \textbf{Terminology and reference resources:} Required use of specific terms, glossaries, or past translations to ensure consistency.

Example: The company previously translated \textit{企業理念} as \textit{Corporate Philosophy}, and this terminology should be maintained consistently across sections and future documents.

\item \textbf{Domain and legal requirements:} Industry-specific norms or legal constraints that affect wording or structure.

Example: The report includes financial statements that must conform to IFRS terminology, and disclosures must reflect the Japan Financial Services Agency's guidelines on non-financial reporting.

\item \textbf{Cultural adaptation:} Modifications made to accommodate cultural expectations or sensitivities.

Example: A phrase like \textit{wa no seishin (和の精神)} may be unfamiliar to global readers and can be replaced with a culturally adapted equivalent such as \textit{a spirit of harmony and mutual respect}.

\item \textbf{Length and formatting:} Constraints or allowances regarding text length, layout, or formatting elements.

Example: Since the English translation must fit within the same layout as the Japanese version, sentence length and paragraph structure must be carefully managed to avoid overflow.

\item \textbf{Localization needs:} Adjustments for language variants, regional conventions, or localized preferences.

Example: Dates, currencies, and units should follow international conventions (e.g., \textit{FY2023} instead of \textit{2023年度}, \textit{million yen} instead of \textit{百万円}), and U.S. spelling is preferred for the target audience.
\end{enumerate}

As these specifications demonstrate, creating appropriate translations requires numerous decisions. 
Defining these requirements in advance helps ensure that translations meet their intended purpose and are suitable for the target context.

\paragraph{Step 2: Define Roles}
Separate tasks performed by the machine, such as generating translations based on the given specifications, from those handled by humans, including supervision, quality assurance, verification of specification adherence, and the management of responsibilities such as meeting deadlines. 
Human reviewers oversee the overall process and focus on elements requiring expert judgment or domain-specific knowledge.

\paragraph{Step 3: Design Instructions Aligned with Specifications}

Create instructions for the LLM that reflect the defined specifications.
Use the listed specifications to identify aspects that can be included in the model instructions.
These instructions may include parameters such as:

\begin{itemize}
    \item Source text information
    \item Target language
    \item Purpose of translation
    \item Target audience
    \item Style, register, and tone
    \item Length and formatting
\end{itemize}

These parameters may vary depending on the specific translation project.
The instructions should also state that LLM must not add any information that is not present in the original text.  
This is especially important in creative translation tasks, where the model may over-generate and introduce information not found in the source text.

Fine-tuning may be applied as needed, such as specifying terminology, aligning with existing translations, using consistent phrasing, maintaining preferred styles, or handling domain-specific vocabulary accurately. 

\paragraph{Step 4: Generate Translation}

Use the instructions to generate the initial translation. 

\paragraph{Step 5: Review and Finalization}

A human reviewer, or a team of reviewers, must carefully check whether the initial translation is accurate and ensure that it follows the defined specifications.  
Any errors should be corrected, and the translation should be finalized for delivery.
This review process should be given sufficient time in the project schedule.  
It is standard practice to have reviewers check the translation to ensure quality and accuracy, even when the initial translation is done by professional human translators.  
The review helps identify issues that a single translator may miss.  
The same applies to LLMs.

\subsection{Human Error Analysis Approach}
\label{appendix:human_error}
Translation evaluation is not always conducted alongside the translation itself.  
There are various situations in which translation evaluation is required, such as deciding whether to accept or reject a translation, comparing multiple translation outputs, or selecting the most suitable version among candidates.  
Evaluation may also be needed for quality control in professional workflows, for assessing MT outputs, or for translator training and certification purposes.

Our error analysis methodology combines ISO 5060, the JTF guidelines, and the MQM framework to create a specification-aware evaluation system.  
This system is designed to score each translation based on how well it meets the predefined specifications.
The more errors are found, the higher the score. 
Therefore, translations of higher quality should receive lower scores.
By using these established frameworks, we develop a practical process that can be applied in various translation contexts.\footnote{Although full-text evaluation is ideal, practical constraints sometimes necessitate the use of samples.
When translation samples are selected for evaluation, ISO 5060, MQM, and the JTF guidelines each offer instructions on how to carry out the sampling process \cite{ISO5060, MQM, JTFevaluation}.}

\paragraph{Step 1: Ensure Specifications Are Accessible}
Before evaluation begins, ensure that the translation specifications are available and accessible.
If, for any reason, clear specifications are not defined at the time of translation, they should be established at this stage.
The specifications should remain accessible throughout the evaluation process so that evaluators can refer to them consistently.
Clear documentation helps ensure that all evaluators apply the same criteria and share a common understanding of the communicative goals of the translation.

\paragraph{Step 2: Define Error Categories Based on Specifications}
Establish error categories aligned with the specifications to ensure consistency.
ISO 5060 defines a translation error as a ``failure to adhere to translation project specifications'' \cite{ISO5060}. 

However, it is important to distinguish between specifications and error categories.
Specifications describe the requirements agreed upon for a project.
In contrast, error categories cover a broader range of issues, including problems that are commonly assumed but not always stated directly in the specifications.
For example, accuracy-related errors or violations of general linguistic conventions, such as grammar or punctuation, are typically included in error taxonomies even when they are not listed in the specifications.
As long as the selected error categories do not contradict the specifications, they can be applied to support consistent evaluation.

Reference standards like the JTF guidelines and MQM for error categorization \cite{JTFevaluation, MQM}. 
MQM provides fine-grained error categories that are especially valuable for detailed error analysis \cite{MQM}.

Error levels can be customized based on project priorities.
For example, when translating a company's corporate philosophy for investor relations materials, where conveying nuance and tone is more important than achieving word-for-word accuracy, the following adjustments can be made:
\begin{itemize} 
\item Accuracy: Prioritize mistranslation errors, while placing less emphasis on over-translation and under-translation. 
\item Style: Use more specific subcategories, such as: 
\begin{itemize} \item Language register: Inappropriate level of formality 
\item Awkward phrasing: Grammatically correct but stylistically poor 
\item Unidiomatic expressions: Unnatural to native speakers 
\item Inconsistent style: Stylistic variations across the document 
\end{itemize} 
\end{itemize}
This targeted approach allows evaluators to focus on errors that are most relevant to the specifications.
Error categories should be adjusted to reflect the specific requirements of each project.

\paragraph{Step 3: Apply Weights to Error Categories}
ISO 5060 and the JTF guidelines apply weights to error categories based on project specifications.  
These weights can be set in advance by the client or project owner.
Since different document types prioritize different aspects of translation quality, stakeholders should agree on appropriate weights \cite{ISO5060, JTFevaluation}:

\begin{itemize}
    \item 2.0 – Highly Important  
    \item 1.5 – Somewhat Important  
    \item 1.0 – Standard Importance (default)  
    \item 0.5 – Less Important  
\end{itemize}

To maintain balance, the average weight across all categories should be approximately 1.0.
The weighting example below places greater emphasis on accuracy and consistent currency formatting, while giving less weight to stylistic elements:\\  

Example: Financial Report---Revenue Forecast Section

\begin{itemize}
    \item Accuracy: 1.0 (Standard): Basic factual correctness is required, with some flexibility in expressing forecasts.
    \item Style: 0.5 (Less Important): A professional tone is preferred, but it has little impact on understanding.
    \item Locale Convention: 1.5 (More Important): All monetary values must be shown in US dollars to ensure a consistent interpretation.
\end{itemize}

\paragraph{Step 4: Select Qualified Evaluators}
Evaluators should be professional translators or subject matter experts who are not only bilingual but also native speakers of the target language.  
Being bilingual alone is not sufficient for proper evaluation; evaluators must have a deep understanding of the linguistic nuances and cultural context of the target language.  

Regarding the qualifications and competencies of evaluators, the ISO 5060 and MQM frameworks provide more rigorous requirements \cite{ISO5060, MQM}.  
Practitioners should refer directly to these standards to ensure that evaluators meet the necessary professional criteria.  

However, these strict requirements can be challenging in practice, as it is often hard to find and recruit qualified evaluators.
When qualified evaluators are difficult to recruit, it may be useful to combine error-based evaluation conducted by available bilingual reviewers with subjective assessment by domain experts in the target language or end users.
Domain experts may detect inconsistencies or errors by closely reading the content, while end users can provide direct feedback on whether the translation feels natural or conveys the intended message.

\paragraph{Step 5: Identify and Assess Errors}
Qualified evaluators identify errors, record them, and assess their severity. 
Severity indicates how serious an error is.  
It should always be judged based on whether the translation achieves its intended purpose and how much real-world impact the error may have \cite{JTFevaluation}.  
If an error has little or no practical impact, assigning a severity level may not be necessary.

Severity levels and their corresponding scores follow \citet{JTFevaluation}:

\begin{itemize}
    \item Neutral (0): No penalty. These include stylistic preferences or repeated minor issues that do not affect comprehension.
    \item Minor (1): Errors that slightly affect readability but do not interfere with understanding.
    \item Major (10): Errors that significantly affect readability and comprehension.
    \item Critical (100): Errors that make the translation unusable and may cause harm, such as health risks, financial losses, or reputational damage. These must be corrected before publication.
\end{itemize}
Whether to count repeated errors multiple times should be decided through agreement among stakeholders \cite{JTFevaluation}.

\paragraph{Step 6: Calculate the Score}
Each identified error receives a score calculated as:
\begin{align}
\text{Error Score} = \text{Category Weight} \times \text{Severity Score}
\end{align}
After assessing all errors, sum the scores to calculate the total error score.  
If severity scoring is not used, simply multiply the number of errors in each category by its assigned weight, and then sum the results.

A lower total score indicates fewer errors, and therefore a higher-quality translation.
These scores make it possible to directly compare different translations.  

For projects that require pass/fail decisions, evaluators can set a passing threshold based on acceptable error levels.  
Refer to the JTF guidelines or the MQM framework for recommended threshold values that fit different translation contexts \cite{JTFevaluation, MQM}.

As the evaluation is based on detailed specifications, it allows for a more objective assessment, reducing subjectivity and ensuring consistent criteria among evaluators.

\subsection{Human Subjective Evaluation Approach}
\label{appendix:human_subjective}
The JTF guidelines emphasize that while error-based evaluation methods provide a systematic approach, they represent only one aspect of translation quality \cite{JTFevaluation}.  
Error-based methods focus primarily on objectively identifiable errors and do not account for subjective quality factors that are essential in certain types of documents, such as advertisements, literary works, corporate vision statements, marketing slogans, and brand messages.
For a more comprehensive assessment of quality, it is advisable to combine error-based evaluation with other approaches, particularly subjective evaluation by experts or end users, depending on the translation context \cite{JTFevaluation}.

Moreover, when qualified evaluators for error-based assessment are unavailable, subjective evaluation can complement error analysis performed by available bilingual reviewers.
Subjective evaluation captures aspects such as clarity, persuasiveness, and appropriateness, which are essential for determining whether a translation effectively serves its intended purpose.

Subjective evaluation can be integrated with error-based assessment in the following ways:

\paragraph{A. Subjective Evaluation by Experts}:
Experts assess translations based on their specialized knowledge and professional judgment.
Examples of such expert evaluations include:
\begin{itemize}
    \item Legal professionals assess the accuracy and appropriateness of legal translations.
    \item Marketing specialists review the effectiveness and cultural relevance of promotional content.
    \item Technical staff on the client side evaluate clarity, precision, and technical correctness in the documentation.
\end{itemize}

\paragraph{B. Subjective Evaluation by End Users}:
End users evaluate translations based on their own perceptions and practical experience.  
For example, in the case of investor relations materials, investors may be asked:
\begin{itemize}
    \item \textit{Did you find the explanation clear, convincing, and appropriate?}
    \item \textit{Were there any unnatural expressions or confusing elements in the translation?}
\end{itemize}

Their feedback is usually collected through surveys or questionnaires and provides valuable insight into how clear and usable the translation is.

Incorporating specifications into translation and evaluation enables both to go beyond basic accuracy and focus on communicative effectiveness.
This ensures translations are not only correct but also appropriate for their intended audiences and contexts.

\section{Why Focus on Japanese-to-English Translation}
\label{appendix:why-JE}
Our experiment focuses on Japanese-to-English translation of integrated reports, which are typically published annually by companies.  
Here, we explain why this particular language pair and translation direction were chosen.

MT research often prioritizes universal approaches, aiming to develop models and evaluation methods that generalize across many language pairs.
For example, \citet{liu-etal-2024-evaluation} note that this focus on generalization is evident in the field's pursuit of standardized methods. 
While such approaches help improve general performance across languages, they may overlook challenges specific to individual language pairs.

Indeed, prior research has reported substantial differences in MT performance between high-resource and low-resource languages \cite{nllbteam2022languageleftbehindscaling, pang-etal-2025-salute}.
These disparities have been attributed not only to the quantity of training data but also to inherent linguistic factors.
For instance, \citet{sarti-etal-2022-divemt} found that post-editing greatly improved English–Italian translations, but had limited impact on English–Turkish and English–Japanese, likely due to word order and morphology differences.
Similarly, \citet{lee-etal-2022-pre} report that mBART \cite{tang-etal-2021-multilingual} performs well across domains but struggles with typologically distant languages, scoring below 3.0 BLEU \cite{papineni-etal-2002-bleu}.

In response, our study takes a more targeted perspective, recognizing that translation difficulty varies widely depending on linguistic distance, grammatical structure, and cultural context.
Japanese–English translation presents unique challenges due to fundamental differences in linguistic structure and writing systems.
Unlike English–European language pairs, which often require minimal restructuring, Japanese–English translation typically involves major changes in sentence structure and word choice \cite{carl-etal-2016-english}.
\citet{ogawa2021difficulty} states that translating between Japanese and English takes significantly more processing time than English–German or English–French translation, leading to a higher cognitive load.

Due to these complexities, general approaches to translation models and evaluation, particularly those designed for multilingual settings, may not fully reflect the specific challenges of Japanese–English translation.
Therefore, a specialized approach is necessary to understand not only how MT systems handle these linguistic difficulties, but also how their outputs can be appropriately evaluated.

We focus on the Japanese-to-English translation direction for two main reasons.
First, ChatGPT and other LLM-based translation systems tend to perform better in English than in many other languages, due to the abundance of high-quality English training data \cite{chowdhery2022palmscalinglanguagemodeling, wendler-etal-2024-llamas}.
For example, OpenAI reports that GPT-3's training data is ``primarily English (93 [percent] by word count),'' and similar English-centric characteristics are reflected in the design and evaluation of GPT-4, as noted in its System Card \cite{brown2020languagemodelsfewshotlearners, openai2024gpt4ocard}.
Working in this direction allows for both practical translation and evaluation while minimizing the influence of data-related limitations.

Second, the demand for Japanese-to-English translation greatly exceeds the supply of qualified human translators.
Japan continues to face a shortage of professionals who can produce high-quality English translations, especially in specialized fields such as investor relations and corporate communications \cite{nagata2025english, ueno2025pronexus}.
Given this shortage, MT guided by translation specifications represents a practical alternative that may help meet the demand for high-quality and cost-effective translations.
In this study, we examine whether this approach can improve translation quality while also addressing the translator shortage and meeting international communication needs.

\begin{table*}[t]
\centering\footnotesize
\begin{tabular}{llc}
\toprule
Industry & Company & Year \\
\midrule
Transportation Equipment & Toyota Motor Corp. & '23 \\
Banks & Mitsubishi UFJ Financial Group, Inc. & '23 \\
Electric Appliances & Sony Group Corp. & '23 \\
Retail Trade & Fast Retailing Co., Ltd. & '23 \\
Services & Recruit Holdings Co., Ltd. & '23 \\
Information \& Communication & Nippon Telegraph and Telephone Corp. & '23 \\
Chemicals & Shin-Etsu Chemical Co., Ltd. & '24 \\
Wholesale Trade & Mitsubishi Corp. & '23 \\
Pharmaceuticals & Chugai Pharmaceutical Co., Ltd. & '23 \\
Other Products & ASICS Corp. & '23 \\
Insurance & Tokio Marine Holdings, Inc. & '23 \\
Foods & Japan Tobacco Inc. & '23 \\
Precision Instruments & Terumo Corp. & '23 \\
Fishery, Agriculture \& Forestry & Nippon Suisan Kaisha, Ltd. & '23 \\
Mining & Japan Petroleum Exploration Co., Ltd. & '23 \\
Construction & Daiwa House Industry Co., Ltd. & '23 \\
Textiles \& Apparels & Goldwin Inc. & '23 \\
Pulp \& Paper & Oji Holdings Corp. & '24 \\
Oil \& Coal Products & ENEOS Holdings, Inc. & '23 \\
Rubber Products & Bridgestone Corp. & '24 \\
Glass \& Ceramics Products & AGC Inc. & '24 \\
Iron \& Steel & Nippon Steel Corp. & '23 \\
Nonferrous Metals & Sumitomo Electric Industries, Ltd. & '23 \\
Metal Products & Sanwa Holdings Corp. & '23 \\
Machinery & Mitsubishi Heavy Industries, Ltd. & '23 \\
Electric Power \& Gas & The Kansai Electric Power Co., Inc. & '23 \\
Land Transportation & Central Japan Railway Co. & '23 \\
Marine Transportation & Nippon Yusen Kabushiki Kaisha & '23 \\
Air Transportation & All Nippon Airways Co., Ltd. & '23 \\
Warehousing \& Harbor Transportation Services & Mitsui-Soko Holdings Co., Ltd. & '23 \\
Securities \& Commodity Futures & Nomura Holdings, Inc. & '24 \\
Other Financing Business & Japan Exchange Group, Inc. & '23 \\
Real Estate & Mitsui Fudosan Co., Ltd. & '23 \\
\bottomrule
\end{tabular}
\caption{Integrated reports of 33 listed companies used in the experiment.}
\label{table:companies}
\end{table*}

\section{Company List}
\label{appendix:company-list}
Table~\ref{table:companies} lists the integrated reports used in this study. 
One company is selected from each of the 33 industry categories defined by the Tokyo Stock Exchange, prioritizing those ranked among the top four in market capitalization within each category during the data collection period (August 28–September 9, 2024).

The table includes the industry, company name, and publication year of the report used. 
Major companies with larger market capitalizations are chosen under the assumption that they are more likely to publish well-developed English versions of their integrated reports. 
As a result, many of the companies listed are well-known Japanese corporations.

\section{Prompt Design for Specification-Aware ChatGPT Translations}
\label{appendix:prompts}
\begin{table*}[t]
    \centering
    \footnotesize
    \renewcommand{\arraystretch}{1}
    \begin{tabular}{p{0.2\linewidth} p{0.7\linewidth}}
        \toprule
        \textbf{Translation Type} & \textbf{Prompt} \\
        \midrule
        ChatGPT basic & Please translate the following Japanese text into English. \\
        \addlinespace[0.5em]
        ChatGPT + Spec & The following Japanese text is an excerpt from the integrated report of [company name], a key part of the company's investor relations materials. Please translate this text into English in a way that will be appealing to international investors. The purpose of this translation is to enhance the company's appeal to a wider audience of investors. Please do not add any additional information. \\
        \addlinespace[0.5em]
        ChatGPT PE + Spec& The following text is a translation of an excerpt from the integrated report of [company name], a key part of the company's investor relations materials. The purpose of this translation is to enhance the company's appeal to a wider audience of investors. The initial translation was done using Google Translate. Please refine this translation to make it more engaging and appealing in English. Please do not add any additional information. \\
        \bottomrule
    \end{tabular}
    \caption{Examples of ChatGPT prompts used in our study.}
    \label{table:prompts}
\end{table*}

Table~\ref{table:prompts} summarizes the prompts used for each ChatGPT translation method.
For ChatGPT basic, ChatGPT receives only a minimal instruction: ``Please translate the following Japanese text into English.''
For ChatGPT + Spec, we incorporate the content of the assumed specifications for the translation of integrated reports:

\begin{itemize} 
\item \textbf{Source text context}: The official company name and a description of the integrated report as an IR document
\item \textbf{Target language}: English
\item \textbf{Intended purpose}: To enhance the company's appeal to a broad audience of investors
\item \textbf{Target audience}: International investors
\item \textbf{Style}: Clear and persuasive, suitable for a global investor audience
\end{itemize}

In the ChatGPT PE + Spec method, the model is instructed to improve the Google Translate output based on the same specifications, without access to the original Japanese text. 
This is a deliberate design choice; providing the source text risks the model disregarding the MT output and producing a new translation from scratch, a phenomenon observed in our initial pilot experiments.
We therefore ensured that the task remained genuine post-editing, focused solely on enhancing the fluency and appeal of the existing translation.
In both specification-aware methods, we instruct ChatGPT to avoid adding information not present in the source.

Since this is not a commissioned project, we define realistic specification parameters based on industry practices.
Items like deadlines and formatting are excluded due to experimental constraints.

\section{Example Comparison of Five Translation Methods: All Nippon Airways Co., Ltd.}
\label{appendix:example_ANA}
\begin{table*}[!t]
\centering\footnotesize\renewcommand{\arraystretch}{1.4}
\begin{tabular}{p{4em} p{38em}}  % 左列を固定幅に変更
\toprule
\textbf{Type} & \textbf{Translation Output} \\ \midrule
\textbf{Source} & ``ワクワク''は, 人を動かすエネルギー. それは人から人へと伝わり、世界をあかるく元気にする。
私たちは, 空の可能性を信じ, ヒト・モノ・コトをつなぐことで, これまでたくさんの笑顔と歓びに出会ってきた. そして今, 新しい未来に向けて, 次のワクワクを生み出す. 
移動をもっと楽しいものにできないか, モノの運び方を新しくできないか, 心躍る日常を届けられないか, どこまでも地球にやさしくなれないか. これまで以上の空の体験とともに, 驚きと感動を創りたい. \\
\textbf{Official} & ``Waku waku'' is what moves people to push what's possible. It's Japanese for the joy and excitement
of discovering the unknown.
And when passed from person to person, becomes a force that creates a brighter world, united in wonder.
The sky is full of possibilities, which has allowed us to bring together people, products and experiences for decades. Now, we're expanding our horizons and spreading ``waku waku'' across the world.
So, we wonder, can we make travel more enjoyable?
Can we renew the way we transport products?
Can we bring excitement to life, while being kind to the earth? Together with an elevated experience in the sky,
we can explore more incredible discoveries that
will lead to a better tomorrow. \\
\textbf{Google} & ``Excitement'' is the energy that moves people. It spreads from person to person, making the world brighter and more energetic. 
We believe in the possibilities of the sky, and by connecting people, things, and events, we have encountered many smiles and joy. And now, we are creating the next excitement for a new future. 
Can we make travel more fun, can we innovate the way we transport things, can we deliver an exciting daily life, and can we be kind to the earth as much as possible? We want to create surprises and excitement along with more exciting sky experiences than ever before.  \\
\textbf{GPT} & ``Excitement'' is the energy that moves people. It spreads from person to person, bringing brightness and vitality to the world.
Believing in the possibilities of the sky, we have connected people, goods, and experiences, encountering countless smiles and joy along the way. Now, we are setting out to create the next source of excitement for a new future.
Can we make travel more enjoyable? Can we reinvent the way goods are transported? Can we deliver everyday moments that make your heart dance? Can we be more environmentally friendly than ever before? Along with new and enhanced experiences in the sky, we aim to create wonder and inspiration. \\
\textbf{GPT+Sp} & Excitement is the energy that moves people. It spreads from person to person, brightening and invigorating the world. 
We believe in the limitless potential of the skies and, through connecting people, goods, and experiences, we have encountered countless smiles and moments of joy. Now, as we look toward a new future, we are ready to create the next wave of excitement.
Can we make travel more enjoyable? Can we innovate the way things are transported? Can we bring a sense of wonder to everyday life? Can we go further in protecting our planet? We aim to create surprises and inspiration with an even greater sky experience. \\
\textbf{GPT PE+Sp} & ``Excitement'' is the spark that moves people, spreading from one person to another, brightening and energizing the world.
Believing in the boundless potential of the skies, we bring people, goods, and experiences together, creating countless moments of joy and countless smiles. Now, we are embarking on a new journey to inspire even greater excitement for the future.
Can we make travel more delightful, revolutionize the way we transport goods, infuse everyday life with excitement, and care for our planet in the best possible ways? We are committed to creating moments of surprise and delight, offering more thrilling experiences in the skies than ever before. \\
\bottomrule
\end{tabular}
\caption{Differences in translations: All Nippon Airways Co., Ltd.}
\label{table:appendix-translations}
\end{table*}

The five translations in Table \ref{table:appendix-translations} illustrate different approaches to conveying the original Japanese text in English.
The source text is an excerpt from the corporate philosophy section of the integrated report published by All Nippon Airways Co., Ltd.

The official translation retains the Japanese term ``waku waku,'' along with an explanatory note.
It is unclear whether the original specifications required preserving the Japanese phrase. 
However, even if the intention was to reflect a sense of Japanese cultural identity, the primary objective should be to ensure that the translation appeals to international investors.
One notable issue is that in the phrase ``[a]nd when passed from person to person, becomes a force,'' the subject is missing, which could be considered an error.
Additionally, the phrase ``explore more incredible discoveries'' may sound awkward, as ``explore'' typically refers to something unknown rather than something already discovered \cite{etymonline_explore}.

Google Translate output is simple and easy to understand, but expressions such as ``make travel more fun'' and ``more exciting sky experiences'' might be too casual and do not fit the context of investor relations materials.
The phrase ``encountered many smiles and joy'' also sounds unnatural, as ``encounter'' is typically used with concrete entities or situations, such as difficulties or opposition, rather than with abstract concepts like joy or smiles \cite{merriamwebster_encounter}.

ChatGPT basic captures the emotional aspects of the original text, particularly with phrases like ``moments that make your heart dance.''
However, the use of ``encountering countless smiles and joy,'' similar to Goolge Translate, does not sound natural in English.

ChatGPT + Spec employs expressions such as ``limitless potential of the skies,'' which are effective in creating a positive and aspirational tone.
However, the use of ``things'' in ``the way things are transported'' sounds casual, and ``even greater sky experience'' would sound more natural if ``experiences'' were used in the plural form.

ChatGPT PE + Spec employs strong and active expressions.
The word ``spark'' in ``[e]xcitement is the spark that moves people'' creates a vivid and powerful impression, while the phrase ``embarking on a new journey to inspire even greater excitement'' conveys a positive and future-oriented feeling.

These observations suggest that each translation method tends to produce distinct results, and that including specifications in ChatGPT prompts may encourage the use of more purposeful and engaging language.

\section{Error Typology for Human Error Evaluations}
\label{appendix:error-typology}

We use the following categories and subtypes for error annotation: Accuracy (subtypes: mistranslation, addition, omission), Linguistic Conventions (grammar, spelling, unintelligible, textual conventions), and Style (language register, awkward style, unidiomatic style, inconsistent style). 
Definitions and examples of each error type are available on the official MQM website (\url{https://themqm.org/downloads/}).
Evaluators are instructed to refer to this table during error analysis. 
However, only the main categories are used for error counting; subtypes are provided to help annotators better understand and identify specific issues.

\section{Syntactic Pattern Analysis of Translation Outputs}
\label{appendix:syntax}

\begin{table*}[t]
\centering \small
\begin{tabular}{lccccc}
\toprule
\textbf{Type} & \textbf{Official} & \textbf{Google} & \textbf{ChatGPT} & \textbf{ChatGPT+Sp} & \textbf{ChatGPT PE+Sp} \\
\midrule
Word Count & 8,776 & \textbf{8,894} & 8,655 & 8,351 & 8,216 \\
Clausal \textit{and}s & 58 & \textbf{66} & 54 & 53 & 57 \\
Rel. Pronouns & 72 & \textbf{83} & 61 & 59 & 61 \\
\textit{and}s / 1000w & 6.61 & \textbf{7.42} & 6.24 & 6.35 & 6.94 \\
RelP / 1000w & 8.20 & \textbf{9.34} & 7.05 & 7.07 & 7.42 \\
\bottomrule
\end{tabular}
\caption{Syntactic feature counts per translation type. RelP = relative pronouns (\textit{which}, \textit{who}, and \textit{that}). Frequencies normalized per 1,000 words.}
\label{table:syntactic_features}
\end{table*}

To investigate stylistic tendencies, we analyzed syntactic patterns across the five translation types, focusing on relative clauses and clausal coordination. Specifically, we counted the number of relative pronouns (\textit{which}, \textit{who}, \textit{that}) and clausal instances of \textit{and} (i.e., those connecting two clauses with subject-verb structures). 
We excluded uses of \textit{that} as complementizers, demonstratives, or in cleft constructions, and excluded \textit{and} used at the phrase or word level.

These structures are common in Japanese texts and may reflect a literal transfer of source syntax. Their frequent use can lead to complex, additive structures that may reduce readability in English.

Table~\ref{table:syntactic_features} reports total word counts, raw counts, and normalized frequencies per 1,000 words. 
The results show that Google Translate and the official translations use relatively more relative clauses and clause-level coordination, suggesting less restructuring. In contrast, ChatGPT outputs display simpler sentence structures regardless of prompt specificity. These patterns indicate that prompt-based LLMs tend to favor fluency and conciseness.

\section{Examples and Error Analysis from Official Translations}
\label{appendix:official_examples}

To explore why the official translations received low ratings, we examined problematic excerpts from companies' corporate philosophies. Table~\ref{table:official_examples} presents these examples.

\begin{table}[t]
\small
\centering
\begin{tabular}{p{0.045\linewidth} p{0.88\linewidth}}
\toprule
\textbf{No.} & \textbf{Excerpt} \\
\midrule
(1) & Opportunities for Life. Faster, simpler and closer to you. Since our foundation, we have connected individuals and businesses, offering both a multitude of choices. (Recruit Holdings Co., Ltd.) \\
\addlinespace[0.5em]
(2) & What we do isn't a job. We enjoy exploring what is possible for our future. We question the status quo, fail well and overcome with resilience. We are a force for change. (Recruit Holdings Co., Ltd.) \\
\addlinespace[0.5em]
(3) & In this era of search, where information has become available anytime anywhere, we need to focus more on proposing the optimal choice. We seek to provide `Opportunities for Life,' much faster, surprisingly simpler and closer than ever before. (Recruit Holdings Co., Ltd.) \\
\addlinespace[0.5em]
(4) & Today, what we mean by Our Hopes for the Future, a world where we are our truest selves, respecting, and inspiring each other. Living together in harmony with our planet---in harmony with People and Nature. (Daiwa House Industry Co., Ltd.)\\
\addlinespace[0.5em]
(5) & Create a virtuous cycle between Society and Earth by fully utilizing less of her limited resources. Make the world a richer, better place by bringing out the best out in people and the potential of buildings. (Daiwa House Industry Co., Ltd.) \\
\addlinespace[0.5em]
(6) & For the sake of the Earth, which future generations of children have entrusted in our care. Together with you. (Bridgestone Corp.) \\
\addlinespace[0.5em]
(7) & The single continuous curve represents the dynamism and our commitment for continuous innovation and delivering value to people and society. (Nippon Telegraph and Telephone Corporation) \\
\bottomrule
\end{tabular}
\caption{Excerpts from the official translation cited in the qualitative analysis.}
\label{table:official_examples}
\end{table}

To begin with, Excerpt (1) contains the phrase ``offering both a multitude of choices,'' in which the use of \textit{both} appears semantically inappropriate.
The word \textit{both} typically introduces two parallel elements, but ``a multitude of choices'' is a singular, collective concept, resulting in a semantic mismatch.
In this context, \textit{both} is presumably intended to refer to the two entities mentioned earlier, \textit{individuals} and \textit{businesses}. 
However, its placement creates a structurally awkward and confusing expression.

Excerpt (2) opens with the sentence ``[w]hat we do isn't a job,'' which appears to aim for an inspirational tone but lacks a clear referent or elaboration.
As a result, its meaning may be ambiguous to readers who are not familiar with the intended message behind the expression.
A clearer alternative might be ``[w]hat we do is more than a job,'' which conveys the intended message more directly.

Excerpt (3) contains the phrase ``in this era of search,'' which is not a commonly used expression in English discourse and may not be immediately clear.
Additionally, the combination of the degree adverb \textit{surprisingly} with the comparative adjective \textit{simpler} creates a stylistic inconsistency.

Excerpt (4) is grammatically incomplete. 
The subject and predicate do not form a complete clause, making the intended meaning difficult to determine.

Excerpt (5) uses the phrase ``fully utilizing less,'' which appears to aim for a concise message about efficiency, likely meaning ``to make the most of fewer resources.'' 
However, the expression is semantically ambiguous. 
The adverb \textit{fully} suggests maximization, while \textit{less} implies minimization, creating a tension that may confuse readers rather than clarify the company's commitment to sustainability.
In addition, the phrase ``bringing out the best out in people'' is grammatically incorrect. 
The structure redundantly includes both ``out'' before and after ``the best,'' where only one instance is appropriate.
A corrected version would be ``bringing out the best in people,'' which is idiomatic and clear.

Excerpt (6) contains grammatical issues.
The phrase ``entrusted in our care'' is unidiomatic. 
In standard English, the verb \textit{entrust} typically appears in the form ``entrust someone with something'' or entrust something to someone.''
In addition, the sentence lacks a main clause and does not constitute a complete grammatical unit.

Finally, Excerpt (7) contains two issues.
First, ``commitment for'' is a grammatical error.
The standard preposition in this context is ``commitment to.''
Second, the coordination of ``continuous innovation,'' a noun phrase, and ``delivering value,'' a gerund phrase, is unbalanced and stylistically awkward.
For clarity and parallel structure, both elements should be in the same grammatical form, such as ``continuous innovation and value creation,'' or ``innovating continuously and delivering value.''

As these examples show, the official translation includes not only grammatical inaccuracies but also semantic and stylistic inconsistencies, which may have contributed to its lower rating in the evaluation.

The characteristics of the source texts themselves may help explain the relatively low ratings of the official translations.
Integrated reports sometimes contain expressions that are abstract, culture-specific, or metaphorical in Japanese, which can result in awkward or even ungrammatical output if translated too literally.
In some cases, the official translations appear to reflect such overly direct translations, suggesting that the translator may have prioritized fidelity to the source text's wording or syntax at the expense of naturalness and clarity in English.
While this approach may have been intentional, for example to preserve a uniquely Japanese tone, it can hinder readability and reduce the overall appeal of the translation.
This is reflected in the lower rankings observed in the subjective evaluation.

However, it is important to note that professional translations are typically produced based on specifications or an internal guideline \cite{ISO17100}.
In practice, translators are expected to follow such instructions from the outset; without them, it would be difficult to even begin the task.
Therefore, the types of problems identified in the official translations, such as grammatical errors or awkward phrasing, are unlikely to stem from missing or unclear specifications.
Rather, these issues may be related to language proficiency or a mismatch between the translator's background and the specific requirements of the task.

Specifications can support translation decisions, but achieving linguistic accuracy and fluency may still require a high level of language proficiency.
As noted earlier (Section \ref{subsection:case_subjective_eval}), human translation quality tends to vary, as translators differ in background and ability \cite{freitag-etal-2023-results, PrietoRamos02042024, VolzThiessen2024}.

Furthermore, as discussed in Appendix~\ref{appendix:why-JE}, Japan continues to face a shortage of translators capable of producing high-quality English translations.
This shortage may have influenced the present results.

In this context, the potential of MT systems that use specifications, such as ChatGPT with customized prompts or post-edited outputs, deserves more attention.
Both ChatGPT + Spec and ChatGPT PE + Spec are favorably evaluated in our study, not only in subjective rankings but also in error-based analysis, suggesting that specification-aware MT may offer a useful complement to traditional workflows.

\end{document}